\documentclass{article}
\usepackage{times}
\usepackage{natbib}

\usepackage[accepted]{icml2014}

% ...
\usepackage[utf8]{inputenc}
\usepackage[english]{babel}
\usepackage{latexsym}
\usepackage{amsthm}
\usepackage{amsmath}
\usepackage{amsfonts}
\usepackage{amssymb}
\usepackage{multirow}
\usepackage{booktabs}
\usepackage{graphicx}
\usepackage{tabularx}
\usepackage{xcolor}
\usepackage{bm}     %for proper bold greek letters

\usepackage{xspace}
\usepackage[compact]{titlesec} %shrinks space around section headings
\usepackage{fixltx2e}

\newcommand{\ICMLpretablevskip}{\vspace{0.1in}}
\newcommand{\ICMLposttablevskip}{\vspace{-0.1in}}
\newcommand{\ICMLprefigvskip}{\vspace{0.1in}}
\newcommand{\ICMLpostfigvskip}{\vspace{-0.13in}}

\newcommand{\ngram}{\mbox{$n$-gram}\xspace}

\newcommand{\unk}{\textsc{Unk}\xspace}

\newcommand{\compose}{\emph{compose}\xspace}

\newcommand{\mb}[1]{\mathbf{#1}}

\newcommand*{\textvector}[1]{\ensuremath{\overrightarrow{\mathrm{#1\vphantom{k}}}}}
\newcommand*{\textvectorsl}[1]{\ensuremath{\overrightarrow{\mathit{#1\vphantom{k}}}}}

\newcommand{\model}[1]{{\small\modelx{#1}}}
\newcommand{\modelx}[1]{\textsf{#1}}
\newcommand{\lblpp}{\model{LBL++}\xspace}
\newcommand{\lblpc}{\model{LBL+c}\xspace}
\newcommand{\lblpo}{\model{LBL+o}\xspace}
\newcommand{\lbl}{\model{LBL}\xspace}

\newcommand{\clbl}{\model{CLBL}\xspace}
\newcommand{\clblpp}{\model{CLBL++}\xspace}

\newcommand{\mkn}{\model{MKN}\xspace}

\newcommand{\Set}[1]{\ensuremath{\mathcal{#1}}}
\newcommand{\V}{\Set{V}\xspace}
\newcommand{\F}{\Set{F}\xspace}

\newcommand{\data}[1]{\textsc{#1}}
\newcommand{\Lcs}{\data{Cs}\xspace}
\newcommand{\Lde}{\data{De}\xspace}
\newcommand{\Les}{\data{Es}\xspace}
\newcommand{\Len}{\data{En}\xspace}
\newcommand{\Lfr}{\data{Fr}\xspace}
\newcommand{\Lru}{\data{Ru}\xspace}

\newcommand{\corpus}[1]{{\small\textsf{#1}}}
\newcommand{\corpusx}[1]{{\footnotesize\textsf{#1}}}
\newcommand{\tool}[1]{\textsl{#1}}
\newcommand{\bleu}{\textsc{Bleu}\xspace}

\newcommand{\BigO}[1]{\ensuremath{\mathcal{O}(#1)}}
\newcommand{\lp}[2]{\data{#1}$\rightarrow$\data{#2}}

\newcommand{\mytilde}{\raise.17ex\hbox{$\scriptstyle\mathtt{\sim}$}}

\newcommand{\rhotimes}{\mbox{$\rho\hspace{-2pt}\times\hspace{-2pt}100$}\xspace}

\newcommand{\newcite}[1]{\citet{#1}}

\newcommand*{\SP}{\ensuremath{\bm{\cdot}}}
\newcommand*{\iteg}[1]{\emph{#1}}  % *i*nline *t*ext *eg* example
\newcommand*{\iten}[1]{(`#1')}     % accompanying English gloss/explanation

\usepackage{hyperref}
  \definecolor{mydarkblue}{rgb}{0,0.08,0.45}
  \hypersetup{ %
    pdftitle={Compositional Morphology for Word Representations and Language Modelling},
    pdfauthor={Jan A. Botha, Phil Blunsom},
    pdfsubject={Proceedings of the International Conference on Machine Learning 2014},
    pdfborder=0 0 0,
    pdfpagemode=UseNone,
    colorlinks=true,
    linkcolor=mydarkblue,
    citecolor=mydarkblue,
    filecolor=mydarkblue,
    urlcolor=mydarkblue,
    pdfview=FitH}

\usepackage{url}
\urldef{\ourcode}{\url}{http://bothameister.github.io}

\begin{document} 

\twocolumn[
\icmltitle{Compositional Morphology for Word Representations and Language Modelling}
\icmlauthor{Jan A.\ Botha}{jan.botha@cs.ox.ac.uk}
\icmlauthor{Phil Blunsom}{phil.blunsom@cs.ox.ac.uk}
\icmladdress{Department of Computer Science,
            University of Oxford, Oxford, OX1 3QD, UK}

% You may provide any keywords that you 
% find helpful for describing your paper; these are used to populate 
% the "keywords" metadata in the PDF but will not be shown in the document
\icmlkeywords{language modelling, morphology, distributed feature representations, machine translation, semantic similarity}

\vskip 0.3in
]

\begin{abstract} 
This paper presents a scalable method for integrating compositional morphological representations into a vector-based probabilistic language model.
Our approach is evaluated in the context of log-bilinear language models, rendered suitably efficient for implementation inside a machine translation decoder by factoring the vocabulary.
We perform both intrinsic and extrinsic evaluations, presenting results on a range of languages which demonstrate that our model learns morphological representations that both perform well on word similarity tasks and lead to substantial reductions in perplexity.
When used for translation into morphologically rich languages with large vocabularies, our models obtain improvements of up to 1.2~\bleu points relative to a baseline system using back-off \ngram models.
\end{abstract} 

\section{Introduction}
The proliferation of word forms in morphologically rich languages presents challenges to the statistical language models (LMs) that play a key role in machine translation and speech recognition.
Conventional back-off \ngram LMs \cite{Chen1998} and the increasingly popular vector-based LMs \cite{Bengio2003,Schwenk2006,Mikolov2010} use parametrisations that do not explicitly encode morphological regularities among related forms, like \iteg{abstract}, \iteg{abstraction} and \iteg{abstracted}.
Such models suffer from data sparsity arising from morphological processes and lack a coherent method of assigning probabilities or representations to unseen word forms.

This work focuses on continuous space language models (CSLMs), an umbrella term for the LMs that represent words with real-valued vectors.
Such word representations have been found to capture some morphological regularity \cite{Mikolov2013a}, but we contend that there is a case for building \emph{a~priori} morphological awareness into the language models' inductive bias.
%it is beneficial to make morphology part of the parametrisation more deeply.
Conversely, compositional vector-space modelling has recently been applied to morphology to good effect \cite{Lazaridou2013,Luong2013}, but lacked the probabilistic \mbox{basis} necessary for use with a machine translation decoder.

The method we propose strikes a balance between probabilistic language modelling and morphology-based representation learning.
Word vectors are composed as a linear function of arbitrary sub-elements of the word, e.g.\ surface form, stem, affixes, or other latent information.
The effect is to tie together the representations of morphologically related words, directly combating data sparsity.
This is executed in the context of a log-bilinear (LBL) LM \cite{Mnih2007}, which is sped up sufficiently by the use of word classing so that we can integrate the model into an open source machine translation decoder\footnote{Our source code for language model training and integration into \tool{cdec} is available from \ourcode} 
and evaluate its impact on translation into 6 languages, including the morphologically complex Czech, German and Russian.

In word similarity rating tasks, our morpheme vectors help improve correlation with human ratings in multiple languages.
Fine-grained analysis is used to determine the origin of our perplexity reductions, while scaling experiments demonstrate tractability on vocabularies of 900k types using 100m+ tokens.

\section{Additive Word Representations}
\label{sec:additive-wordreps-idea}
A generic CSLM associates with each word type $v$ in the vocabulary $\V$ a $d$-dimensional feature vector $\mathbf{r}_v \in \mathbb{R}^d$.
Regularities among words are captured in an opaque way through the interaction of these feature values and a set of transformation weights.
This leverages linguistic intuitions only in an extremely rudimentary way, in contrast to hand-engineered linguistic features that target very specific phenomena, as often used in supervised-learning settings.

We seek a compromise that retains the unsupervised nature of CSLM feature vectors, but also incorporates \emph{a~priori} linguistic knowledge in a flexible and efficient manner.
In particular, \emph{morphologically related words should share statistical strength} in spite of differences in surface form. 

To achieve this, we define a mapping $\mu:\V \mapsto \F^+$ of a surface word into a variable-length sequence of \emph{factors}, i.e.\ $\mu(v) = (f_1, \dots, f_K)$, where $v\in \V$ and $f_i \in \F$.
Each factor~$f$ 
has an associated \emph{factor feature vector} $\mathbf{r}_f \in \mathbb{R}^d$.
We thereby factorise a word into its surface morphemes, although the approach could also incorporate other information, e.g.\ lemma, part of speech.

The vector representation $\tilde{\mb{r}}_v$ of a word $v$ is computed as a function $\omega_\mu(v)$ of its factor vectors. We use addition as composition function: %
$\tilde{\mb{r}}_v =\omega_\mu(v)= \sum_{f \in \mu(v)} \mathbf{r}_f$.
The vectors of morphologically related words become linked through shared factor vectors
(notation: \textvector{word}, \textvectorsl{factor}),
\begin{center}\vspace{-6pt} %
\begin{tabular}{rcl} %
\textvector{imperfection} &$=$& $\textvectorsl{im} + \textvectorsl{perfect} + \textvectorsl{ion}$ \\
\textvector{perfectly} &$=$& $\textvectorsl{perfect} + \textvectorsl{ly}$.
\end{tabular}  
\end{center} %
\vspace{-6pt}%

Furthermore, representations for out-of-vocabulary (OOV) words can be constructed using their available morpheme vectors.

We include the surface form of a word as a factor itself.
This accounts for noncompositional constructions ($\textvector{greenhouse} = \textvectorsl{greenhouse} + \textvectorsl{green} + \textvectorsl{house}$), and makes the approach more robust to noisy morphological segmentation.
This strategy also overcomes the order-invariance of additive composition 
($\textvector{hangover} \neq \textvector{overhang}$).

The number of factors per word is free to vary over the vocabulary, making the approach applicable across the spectrum of more fusional languages (e.g.\ Czech, Russian) to more agglutinative languages (e.g.\ Turkish).
This is in contrast to \emph{factored language models} \cite{Alexandrescu2006}, which assume a fixed number of factors per word. 
Their method of concatenating factor vectors to obtain a single representation vector for a word can be seen as enforcing a partition on the feature space.
Our method of addition avoids such a partitioning and better reflects the absence of a strong intuition about what an appropriate partitioning might be.
A limitation of our method compared to theirs is that the deterministic mapping $\mu$ currently enforces a single factorisation per word type, which sacrifices information obtainable from context-disambiguated morphological analyses.

Our additive composition function can be regarded as an instantiation of the weighted addition strategy that performed well in
a distributional compositional approach to derivational morphology \cite{Lazaridou2013}.
Unlike the recursive neural-network method of \newcite{Luong2013}, we do not  impose a single tree structure over a word, which would ignore the ambiguity inherent in words like un[[lock]able] vs.\ [un[lock]]able. 
In contrast to these two previous approaches to morphological modelling, our additive representations are readily implementable in a probabilistic language model suitable for use in a decoder.

\section{Log-Bilinear Language Models}
Log-bilinear (LBL) models \cite{Mnih2007} are an instance of CSLMs that
make the same Markov assumption as \ngram language models. The probability of a sentence $\mb{w}$ is 
decomposed over its words, each conditioned on the \mbox{$n$--1}~preceding words: %
\mbox{%
    $ %
    P(\mb{w}) \approx \prod_i P\left(w_i | w_{i-n+1}^{i-1}\right)
    $}. %
These distributions are modelled by a smooth scoring function $\nu(\cdot)$ over vector representations of words. 
In contrast, discrete \ngram models are estimated by smoothing and backing off over empirical distributions \cite{Kneser1995,Chen1998}.

The LBL predicts the vector $\mb{p}$ for the next word as a function of the context vectors $\mb{q}_j \in \mathbb{R}^d$ of the preceding words, 
\begin{equation} %%
\mb{p} = \sum_{j=1}^{n-1}  \mb{q}_j C_j,
\end{equation} % %
where $C_j \in \mathbb{R}^{d\times d}$ are position-specific transformations.

$\nu(w)$ expresses how well the observed word $w$ fits that prediction and is defined as $\nu(w)=\mb{p} \cdot \mb{r}_w + b_w$,
where $b_w$ is a bias term encoding the prior probability of a word type.
Softmax then yields the word probability as %
\begin{equation} %
P\left(w_i | w_{i-n+1}^{i-1}\right) = 
\frac{\exp\left(  \nu(w_i) \right) }
{\sum_{v \in \V} \exp\left( \nu(v)\right)}.
\end{equation} %
This model is subsequently denoted as {\bf \lbl} with parameters
\mbox{$\Theta_{\lbl}=\left(C_j,Q,R,\mb{b} \right)$}, where $Q, R \in \mathbb{R}^{|\V|\times d}$ contain the word representation vectors as rows, and \mbox{$\mb{b} \in \mathbb{R}^{|\V|}$}.
$Q$ and $R$ imply that separate representations are used for conditioning and output.

\subsection{Additive Log-Bilinear Model}
\label{sec:additive-lbl}
We introduce a variant of the LBL that makes use of additive representations (\S\ref{sec:additive-wordreps-idea}{}) by associating 
the \emph{composed word vectors}  $\tilde{\mb{r}}$ and $\tilde{\mb{q}}_j$ with the target and context words, respectively.
The representation matrices $Q^{(f)}, R^{(f)}\in \mathbb{R}^{|\F|\times d}$ thus contain a vector for each factor type. This model is designated {\bf \lblpp} and has parameters
\mbox{$\Theta_{\lblpp}=\left(C_j,Q^{(f)}, R^{(f)},\mb{b} \right)$}.
Words sharing factors are tied together, which is expected to improve performance on rare word forms.

Representing the mapping $\mu$ with a sparse transformation matrix $M\in \mathbb{Z}_{+}^{\V \times |\F|}$,
where a row vector $\mb{m}_v$ has some non-zero elements to select factor vectors,
establishes the relation between word and factor representation matrices as $R = MR^{(f)}$ and $Q = MQ^{(f)}$.
In practice, we exploit this for test-time efficiency---word vectors are compiled offline so that the computational cost of \lblpp probability lookups is the same as for the \lbl.

We consider two obvious variations of the \lblpp to evaluate the extent to which interactions between context and target factors affect the model: {\bf \lblpo} only factorises {\bf o}utput words and retains simple word vectors for the context (i.e. $Q\equiv Q^{(f)}$), while {\bf \lblpc} does the reverse, only factorising {\bf c}ontext words.\footnote{The +c, +o and ++ naming suffixes denote these same distinctions when used with the \clbl model introduced later.}
Both reduce to the \lbl when setting $\mu$ to be the identity function, such that $\V\equiv\F$.

The factorisation permits an approach to unknown context words that is less harsh than the standard method of replacing them with a global unknown symbol---instead, a vector can be constructed from the known factors of the word (e.g.\ the observed stem of an unobserved inflected form).
A similar scheme can be used for scoring unknown target words, but requires changing the event space of the probabilistic model.
We use this vocabulary stretching capability in our word similarity experiments, but leave the extensions for test-time language model predictions as future work.

\begin{figure}[tb]
    \ICMLprefigvskip
    \centering
    \includegraphics[width=0.48\textwidth]{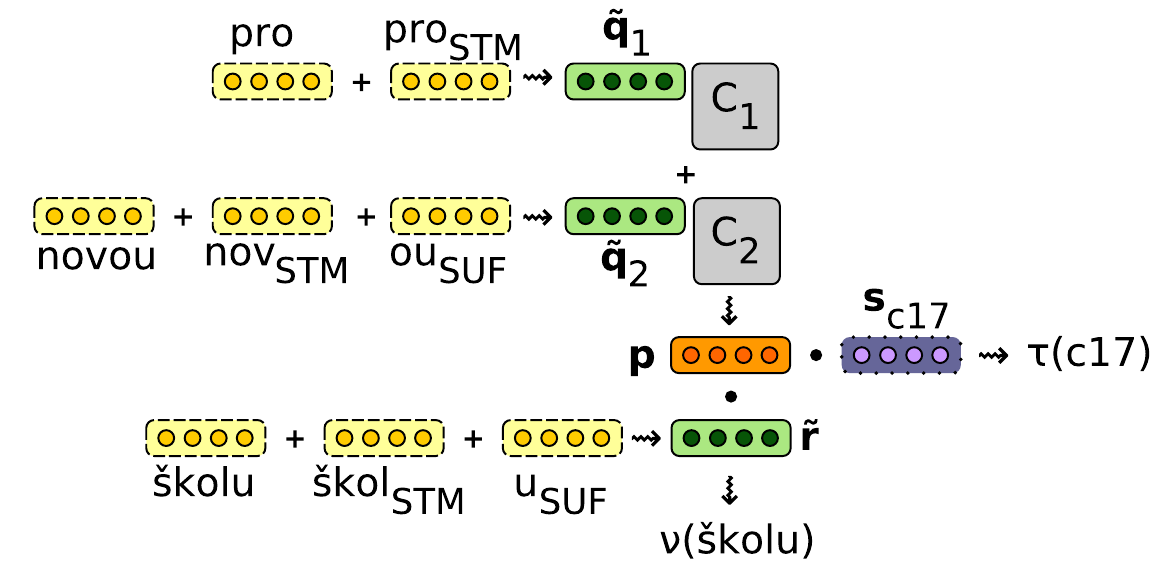}
    \caption{\textbf{Model diagram.} Illustration of how a \mbox{$3$-gram} \clblpp model treats the Czech phrase \iteg{pro novou školu} \iten{for the new school}, 
      assuming the target word \iteg{školu} is clustered into word class~17 by the method described in \S\ref{sec:class-based}.
    }
    \label{fig:instance}
    \ICMLpostfigvskip
\end{figure}

\subsection{Class-based Model Decomposition}
\label{sec:class-based}
The key obstacle to using CSLMs in a decoder is the expensive normalisation over the vocabulary. 
Our approach to reducing the computational cost of normalisation is to use a class-based decomposition of the probabilistic model \cite{Goodman2001a,Mikolov2011}.
Using Brown-clustering \cite{Brown1992},\footnote{In preliminary experiments, Brown clusters gave better perplexities than frequency-binning \cite{Mikolov2011}.}
 we partition the vocabulary into 
$|\Set{C}|$ classes, denoting as $\Set{C}_c$ the set of vocabulary items in class $c$, such that $\V = \Set{C}_1\cup\dots\cup\Set{C}_{|\Set{C}|}$.

In this model, the probability of a word conditioned on the history $h$ of $n-1$ preceding words
is decomposed as
\begin{equation} %
P(w|h) = P(c|h) P(w|h,c).
\end{equation} %
This class-based model, {\bf \clbl}, extends over the LBL by associating a representation vector $\mb{s}_c$ and bias parameter $t_c$ to each class $c$, such that
\mbox{$\Theta_{\clbl}=\left(C_j,Q,R,S,\mb{b},\mb{t} \right)$}.
The same prediction vector $\mb{p}$ is used to compute both class score $\tau(c)=\mb{p}\cdot\mb{s}_c+t_c$ and word score $\nu(w)$, which are normalised separately:
\begin{align} %
P(c | h) &= 
\frac{\exp\left(\tau(c)\right) }
{\sum_{c'=1}^{|\Set{C}|} \exp\left(\tau(c')\right)}
\\
P(w | h,c) &= 
\frac{\exp\left( \nu(w) \right) }
{\sum_{v'\in \Set{C}_c} \exp\left(\nu(v')\right)}.
\end{align} %

We favour this flat vocabulary partitioning for its computational adequacy, simplicity and robustness.
Computational adequacy is obtained by using $|\Set{C}| \approx |\V|^{0.5}$, thereby reducing the $\BigO{|\V|}$ normalisation operation of the \lbl to two $\BigO{|\V|^{0.5}}$ operations in the \clbl.

Other methods for achieving more drastic complexity reductions exist in the form of frequency-based truncation, shortlists \cite{Schwenk2004}, or casting the vocabulary as a full hierarchy \cite{Mnih2008} or partial hierarchy \cite{Le2011}.
We expect these approaches could have adverse effects in the rich morphology setting, where much of the vocabulary is in the long tail of the word distribution.

\subsection{Training \& Initialisation}
Model parameters $\Theta$ are estimated by optimising an L2-regularised log likelihood objective.
Training the \clbl and its additive variants directly against this objective is fast because normalisation of model scores, which is required in computing gradients, is over a small number of events. % (classes, and word types in a given class).

For the classless LBLs we use noise-contrastive estimation (NCE) \cite{Gutmann2012,Mnih2012} to avoid normalisation during training. This leaves the expensive test-time normalisation of LBLs unchanged, precluding their usage during decoding.

Bias terms $\mb{b}$ (resp.\ $\mb{t}$) are initialised to the log unigram probabilities of words (resp.\ classes) in the training corpus, with Laplace smoothing, while all other parameters are initialised randomly according to sharp, zero-mean Gaussians.
Representations are thus learnt from scratch and not based on publicly available embeddings,
meaning our approach can easily be applied to many languages.

Optimisation is performed by stochastic gradient descent with updates after each mini-batch of $L$ training examples.
We apply AdaGrad \cite{Duchi2011} and tune the step-size $\xi$ on development data.\footnote{%
    $L$=10k--40k, $\xi$=0.05--0.08, dependent on $|\V|$ and data size.
}\ %
We halt training once the perplexity on the development data starts to increase. 

\section{Experiments}
The overarching aim of our evaluation is to investigate the effect of using the proposed additive representations across languages with a range of morphological complexity.

Our intrinsic language model evaluation has two parts.
We first perform a model selection experiment on small data to consider the relative merits of using additive representations for context words, target words, or both, and to validate the use of the class-based decomposition.

Then we consider class-based additive models trained on tens of millions of tokens and large vocabularies.
These larger language models are applied in two extrinsic tasks: i)~a word-similarity rating experiment on multiple languages, aiming to gauge the quality of the induced word and morpheme representation vectors; ii)~a machine translation experiment, where we are specifically interested in testing the impact of an LBL LM feature when translating into morphologically rich languages.

\begin{table}[tb]%
    \caption{\textbf{Corpus statistics.} The number of sentence pairs for a row \data{X} refers to the English$\rightarrow$X parallel data (but row \Len has Czech as source language).}
    \label{tbl:corpus_stats}
    \ICMLpretablevskip
    \centering
    \begin{small}
        \begin{tabular}{l*{2}{r}|*{2}{r}c} \toprule
            & \multicolumn{2}{c}{\textsc{Data-1m}}
            & \multicolumn{3}{c}{\textsc{Data-Main}} 
            \\
            & \multicolumn{1}{c}{Toks.}  & \multicolumn{1}{c}{$|\V|$}  
            & \multicolumn{1}{c}{Toks.}  & \multicolumn{1}{c}{$|\V|$} 
            & Sent. Pairs
            \\ \midrule
            %%%%%% 'mono' data; 
            \Lcs & 1m & 46k & 16.8m & 206k & 0.7m \\
            \Lde & 1m & 36k & 50.9m & 339k & 1.9m \\
            \Len & 1m & 17k & 19.5m & 60k  & 0.7m \\
            \Les & 1m & 27k & 56.2m & 152k & 2.0m \\
            \Lfr & 1m & 25k & 57.4m & 137k & 2.0m \\
            \Lru & 1m & 62k & 25.1m & 497k & 1.5m \\ \bottomrule
        \end{tabular}
    \end{small}
    \ICMLposttablevskip
\end{table}

\subsection{Data \& Methods}\label{sec:methods}
We make use of data from the 2013 ACL Workshop on Machine Translation.\footnote{\url{http://www.statmt.org/wmt13/translation-task.html}}\ %
We first describe data used for translation experiments, since the monolingual datasets used for language model training were derived from that. The language pairs are 
English$\rightarrow$\{German, French, Spanish, Russian\} and English$\leftrightarrow$Czech.
Our parallel data comprised the \corpus{Europarl-v7} and \corpus{news-commentary} corpora, except for English--Russian where we used \corpus{news-commentary} and the \corpus{Yandex} parallel corpus.\footnote{\url{https://translate.yandex.ru/corpus?lang=en}}
Pre-processing involved lowercasing, tokenising and filtering to exclude sentences of more than 80 tokens or substantially different lengths.

$4$-gram language models were trained on the target data in two batches: 
\textsc{Data-1m} consists of the first million tokens only, while \textsc{Data-Main} is the full target-side data.
Statistics are given in \autoref{tbl:corpus_stats}.
\corpus{newstest2011} was used as development data\footnote{For Russian, some training data was held out for tuning.}
for tuning language model hyperparameters,
while intrinsic LM evaluation was done on
\corpus{newstest2012}. As metric, we use model perplexity (PPL)
$\mathrm{exp}(-\frac{1}{N}\sum_{i=1}^N \mathrm{ln}\, P(w_i))$, where $N$ is the number of test tokens.
In addition to contrasting the LBL variants, we also use modified Kneser-Ney \ngram models (\mkn{s}) \cite{Chen1998} as baselines.

\paragraph{Language Model Vocabularies.}
Additive representations that link morphologically related words specifically aim to improve modelling of the long tail of the lexicon, so we do not want to prune away all rare words, as is common practice in language modelling and word embedding learning.
We define a \emph{singleton pruning rate} $\kappa$, and randomly replace that fraction of words occurring only once in the training data with a global \unk symbol.
$\kappa=1$ would imply a unigram count cut-off threshold of 1.
Instead, we use low pruning rates\footnote{ %
    \textsc{Data-1m}: $\kappa=0.2$;
    \textsc{Data-Main}: $\kappa=0.05$
} %
and thus model large vocabularies.\footnote{ %
    We also mapped digits to~0, 
    and cleaned the Russian data by replacing tokens having \textless80\% Cyrillic characters 
    with \unk.
}

\paragraph{Word Factorisation $\mu$.}
We obtain labelled morphological segmentations from the \mbox{unsupervised} segmentor \tool{Morfessor Cat-MAP} \cite{Creutz2007}.
The mapping $\mu$ of a word is taken as its surface form and the morphemes identified by Morfessor.
Keeping the morpheme labels 
allows the model to learn separate vectors for, say, \iteg{in\textsuperscript{stem}} the preposition and \iteg{in\textsuperscript{prefix}} occurring as \iteg{in\SP{}appropriate}.
By not post-processing segmentations in a more sophisticated way, 
we keep the overall method more language independent.

\subsection{Intrinsic Language Model Evaluation}
\label{sec:eval-lm}

\begin{figure}[tb]
    \ICMLprefigvskip
    \centering 
    \includegraphics[ width=0.44\textwidth]{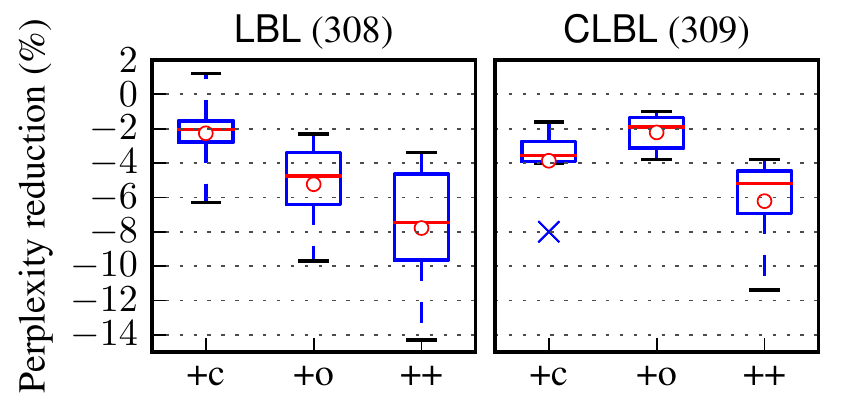}
    \caption{\textbf{Model selection results.} Box-plots show the spread, across 6 languages, of relative perplexity reductions obtained by each type of additive model against its non-additive baseline, for which  median absolute perplexity is given in parentheses; for \mkn, that is 348.
        Each box-plot summarises the behaviour of a model across languages. Circles give sample means, while crosses show outliers beyond 3$\times$ the inter-quartile range.
    }
    \label{fig:boxplots-ppl-1m}
    \ICMLpostfigvskip
\end{figure}

\paragraph{Results on \textsc{Data-1m}.}
The use of morphology-based, additive representations for both context and output words (models++) yielded perplexity reductions on all 6 languages when using 1m training tokens. Furthermore, these double-additive models consistently outperform the ones that factorise only context (+c) or only output (+o) words, indicating that context and output contribute complementary information and supporting our hypothesis that is it beneficial to model morphological dependencies across words. The results are summarised in \autoref{fig:boxplots-ppl-1m}.

For lack of space we do not present numbers for individual languages, but report that the impact of \clblpp varies by language, correlating with vocabulary size:
Russian benefited most, followed by Czech and German.  
Even on English, often regarded as having simple morphology, the relative improvement is 4\%.

The relative merits of the +c and +o schemes depend on which model is used as starting point.
With \lbl, the output-additive scheme (\lblpo) gives larger improvements than the context-additive scheme (\lblpc). The reverse is true for \clbl, indicating the class decomposition dampens the effectiveness of using morphological information in output words. 

The use of classes increases perplexity slightly compared to the \lbl{}s, but this is in exchange for much faster computation of language model probabilities, allowing the \clbl{}s to be used in a machine translation decoder (\S\ref{sec:eval-mt}).

% % % % % LM PPL TABLES % % % % % % % % % % % %
\begin{table}[tb]
    \caption{ %
      \textbf{Test-set perplexities on \textsc{Data-Main}} using two vocabulary pruning settings.
        Percentage reductions are relative to the preceding model, 
        e.g.\ the first Czech \modelx{CLBL} improves over \modelx{MKN} by 20.8\% (Rel.1); the \modelx{CLBL++} improves over that \modelx{CLBL} by a \mbox{\emph{further}} 5.9\% (Rel.2).
    }
    \label{tbl:ppl_mono}
    \ICMLpretablevskip
    \centering     %
    \begin{small}
    % % % % % Table data for perplexity on 'mono' training tokens
% %
\begin{tabular}{l@{}>{\quad}  lr r r<{\%}  r r<{\%}  } \toprule
& & \mkn  & \multicolumn{2}{c}{\clbl} &\multicolumn{2}{c}{\clblpp} \\ 
\cmidrule(lr){3-3} \cmidrule(lr){4-5} \cmidrule(lr){6-7}
& & \multicolumn{1}{r}{PPL} & \multicolumn{1}{r}{PPL} & \multicolumn{1}{c}{Rel.1}
       & \multicolumn{1}{r}{PPL} & \multicolumn{1}{c}{Rel.2} \\
\midrule
\underline{$\kappa$=0.05}  
% % % finaltest (newstest2012) scores
& \Lcs & 862 & 683 & -20.8 & 643 & -5.9 \\ % & -25.4 (commented are the direct %decrease of clbl++ vs mkn)
& \Lde & 463 & 422 & -8.9  & 404 & -4.2 \\ % & -12.7
& \Len & 291 & 281 & -3.4  & 273 & -2.8 \\ % & -6.2
& \Les & 219 & 207 & -5.7  & 203 & -1.9 \\ % & -7.5
& \Lfr & 243 & 232 & -4.9  & 227 & -1.9 \\ % & -6.7
& \Lru & 390 & 313 & -19.7 & 300 & -4.2 \\ % & -23.1
\midrule
\underline{$\kappa$=1.0}  
% % % finaltest (newstest2012) scores
& \Lcs & 634 & 477 & -24.8 & 462 & -3.1 \\ %& -27.1  (commented are the direct %decrease of clbl++ vs mkn)
& \Lde & 379 & 331 & -12.6 & 329 & -0.9 \\ %& -13.4
& \Len & 254 & 234 & -7.6  & 233 & -0.7     \\ %& 
& \Les & 195 & 180 & -7.7  & 180 & 0.02 \\ %& -7.7
& \Lfr & 218 & 201 & -7.7  & 198 & -1.3 \\ %& -8.9
& \Lru & 347 & 271 & -21.8 & 262 & -3.4 \\ %& -24.5
\bottomrule   
\end{tabular}

    \end{small}
    \ICMLposttablevskip
\end{table}
% % % % % % % % % % % % % % % % % % % % % % %

\paragraph{Results on \textsc{Data-Main}.}
Based on the outcomes of the small-scale evaluation, we focus our main language model evaluation on the additive class-based model \clblpp in comparison to \clbl and \mkn baselines, using the larger training dataset, with vocabularies of up to 500k types. 

The overall trend that morphology-based additive representations yield lower perplexity carries over to this larger data setting, again with the biggest impact being on Czech and Russian (\autoref{tbl:ppl_mono}, top). Improvements are in the \mbox{2\%--6\%} range, slightly lower than the corresponding differences on the small data.

Our hypothesis is that the much of the improvement is due to the additive representations being especially beneficial for modelling rare words. We test this by repeating the experiment under the condition where all word types occurring only once are excluded from the vocabulary \mbox{($\kappa$=1)}. If the additive representations were not beneficial to rare words, the outcome should remain the same. Instead, we find the relative improvements become a lot smaller (\autoref{tbl:ppl_mono}, \mbox{bottom}) than when only excluding some singletons (\mbox{$\kappa$=0.05}), which supports that hypothesis.

\paragraph{Analysis.}
Model perplexity on a whole dataset is a convenient summary of its intrinsic performance, but such a global view does not give much insight into \emph{how} one model outperforms another.
We now partition the test data into subsets of interest and measure PPL over these subsets.

\begin{figure}[t]
    \ICMLprefigvskip
    \centering 
    %trim left bottom right top
    \includegraphics[trim=0 0mm 0 0,clip, width=0.46\textwidth]{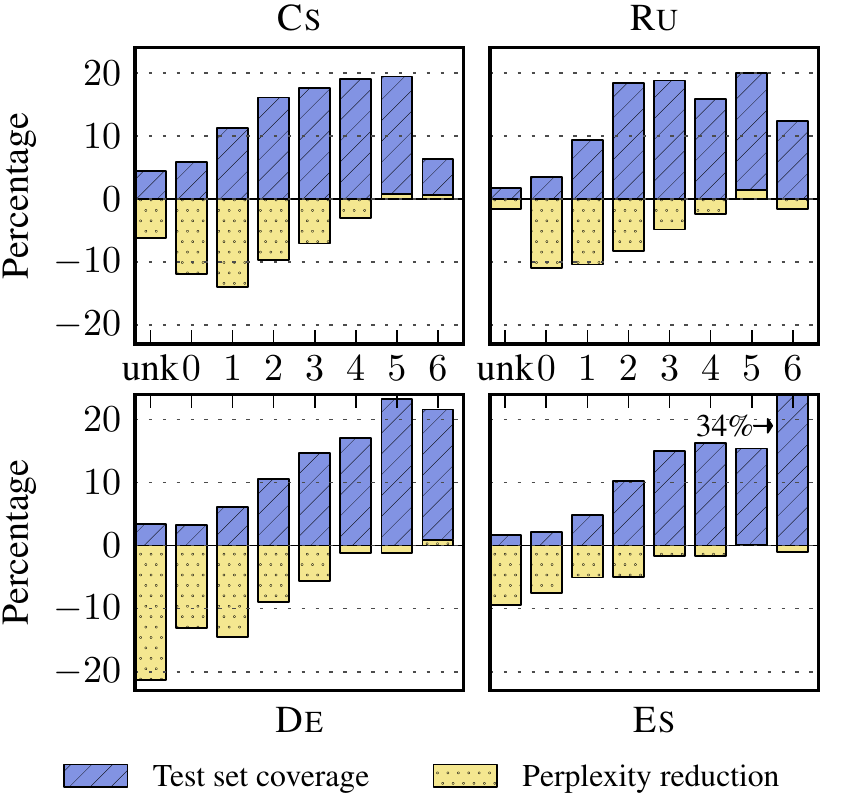}
    \caption{\textbf{Perplexity reductions by token frequency,} \modelx{CLBL++} relative to \modelx{CLBL}.
        Dotted bars extending further down are better.
        A bin labelled with a number $x$ contains those test tokens that occur 
        $y \in [10^x,10^{x+1})$ times in the training data.
        Striped bars show percentage of test-set covered by each bin.
                }
    \label{fig:ppl-by-freq}
\end{figure}
\begin{figure}[t]
  \ICMLprefigvskip
    \centering 
    \includegraphics[ %trim=0 1mm 1mm 1mm,clip,
     width=0.44\textwidth]{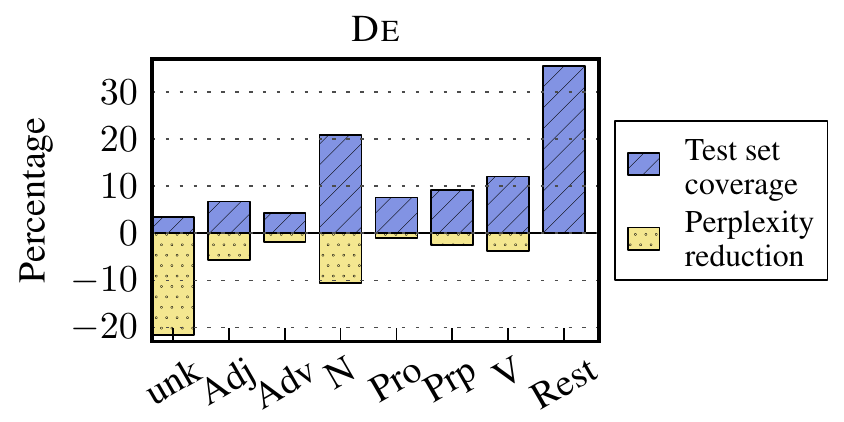}
     \caption{\textbf{Perplexity reductions by part of speech,} \modelx{CLBL++} relative to \modelx{CLBL} on German.
        Dotted bars extending further down are better.
        Tokens tagged as foreign words or other opaque symbols resort under ``Rest''.
        Striped bars as in \autoref{fig:ppl-by-freq}
              }
    \label{fig:ppl-de-by-pos}\vspace{-0.5pt}
\end{figure}

We first partition on token frequency, as computed on the training data. \autoref{fig:ppl-by-freq} provides further evidence that the additive models have most impact on rare words generally, and not only on singletons.
Czech, German and Russian see relative PPL reductions of 8\%--21\% for words occurring fewer than 100 times in the training data.
Reductions become negligible for the high-frequency tokens. These tend to be punctuation and closed-class words, where any putative relevance of morphology is overwhelmed by the fact that the predictive uncertainty is very low to begin with (absolute PPL\textless10 for the highest frequency subset).
For the morphologically simpler Spanish case, PPL reductions are generally smaller across frequency scales.

We also break down PPL reductions by part of speech tags, focusing on German. 
We used the decision tree-based tagger of \newcite{Schmid2008}.
Aside from unseen tokens, the biggest improvements are on nouns and adjectives (\autoref{fig:ppl-de-by-pos}),
suggesting our segmentation-based representations help abstract over German's productive compounding.

German noun phrases require agreement in \mbox{gender}, case and number, which are marked overtly with fusional morphemes, and we see large gains on such test \ngram{}s:
15\% improvement on adjective-noun sequences,
and 21\% when considering the more specific case of adjective-adjective-noun sequences.
An example of the latter kind is 
\iteg{der 
  ehemalig\SP{}e 
  sozial\SP{}ist\SP{}isch\SP{}e
  bildung\SP{}s\SP{}minister}
\iten{the former socialist minister of education}, where the morphological agreement surfaces in the \mbox{repeated e-suffix}.

\begin{figure}[tb]
\ICMLprefigvskip
    \centering 
    \includegraphics[width=0.38\textwidth]{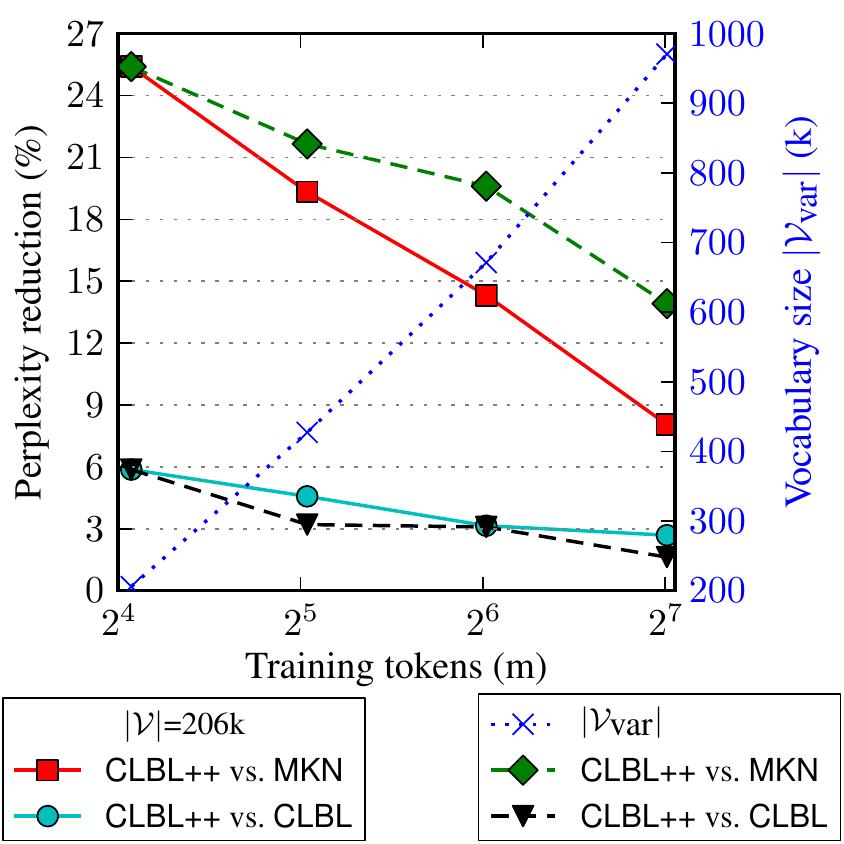}
    \caption{\textbf{Scaling experiment.} Relative perplexity reductions obtained when varying the Czech training data size (16m--128m).
        In the first setting, the vocabulary was held fixed as data size increased($|\V|$);
        in the second it varied freely across sizes ($|\V_{\textbf{var}}|$). % (diamonds, triangles).
    }
    \label{fig:datascale}
\ICMLpostfigvskip
\end{figure}

We conducted a final scaling experiment on Czech by training models on increasing amounts of data from the monolingual news corpora.
Improvements over the \mkn baseline decrease, but remain substantial at 14\% for the largest setting when allowing the vocabulary to grow with the data.
Maintaining a constant advantage over \mkn requires also increasing the dimensionality $d$ of representations \cite{Mikolov2013}, but this was outside the scope of our experiment.
Although gains from the additive representations over the \clbl diminish down to \mbox{2\%--3\%} at the scale of 128m training tokens (\autoref{fig:datascale}),
these results demonstrate the tractability of our approach on very large vocabularies of nearly 1m types.

\subsection{Task 1: Word Similarity Rating}
In the previous section, we established the positive role that morphological awareness played in building continuous-space language models that better predict unseen text.
Here we focus on the quality of the word representations learnt in the process.
We evaluate on a standard word similarity rating task, where one measures the correlation between cosine-similarity scores for pairs of word vectors and a set of human similarity ratings.
An important aspect of our evaluation is to measure performance on multiple languages using a single unsupervised, model-based approach.

Morpheme vectors from the \clblpp enable handling OOV test words in a more nuanced way than using the global unknown word vector. 
In general, we compose a vector $\tilde{\mb{u}}_v=[\tilde{\mb{q}}_v;\tilde{\mb{r}}_v]$ for a word $v$
according to a \emph{post hoc} word map $\mu'$ by summing and concatenating     the factor vectors $\mb{r}_f$ and $\mb{q}_f$, where \mbox{$f \in \mu'(v) \cap \F$}.
This ignores unknown morphemes occurring in OOV words, and uses  $[\mb{q}_{\text{\unk}};\mb{r}_{\text{\unk}}]$ for $\tilde{\mb{u}}_{\text{\unk}}$ only if all morphemes are unknown. 

To see whether the morphological representations improve the quality of vectors for known words,
we also report the correlations obtained when using the \clblpp word vectors directly, resorting to $\tilde{\mb{u}}_{\text{\unk}}$ for all OOV words \mbox{$v\notin \V$} (denoted ``$-$\compose'' in the results).
This is also the strategy that the baseline \clbl model is forced to follow for OOVs.

We evaluate first using the English rare-word dataset (\data{Rw}) created by \newcite{Luong2013}.
Its 2034 word pairs contain more morphological complexity than other well-established word similarity datasets, e.g.\ crudeness---impoliteness.
We compare against their context-sensitive morphological recursive neural network (csmRNN), using Spearman's rank correlation coefficient, $\rho$.
\autoref{tbl:wordsim_rw} shows our model obtaining a $\rho$-value slightly below the best csmRNN result,
but outperforming the csmRNN that used an alternative set of embeddings for initialisation.

This is a strong result given that our vectors come from a simple linear probabilistic model that is also suitable for integration directly into a decoder for translation (\S\ref{sec:eval-mt}) or speech recognition, which is not the case for csmRNNs.
Moreover, the csmRNNs were initialised with high-quality, publicly available word embeddings trained over weeks on much larger corpora of 630--990m words \cite{CollobertWeston2008,HSMN2012},
in contrast to ours that are trained from scratch on much less data.
This renders our method directly applicable to languages which may not yet have those resources.

Relative to the \clbl baseline, our method performs well on datasets across four languages.
For the English \data{Rw}, which was designed with morphology in mind, the gain is 64\%.
But also on the standard English WS353 dataset \cite{Finkelstein2002}, we get a 26\% better correlation with the human ratings.
On German, the \clblpp obtains correlations up to three times stronger than the baseline, and 39\% better for French (\autoref{tbl:wordsim_multilang}).

\begin{table}[tb]
    \caption{\textbf{Word-pair similarity task.} Spearman's \rhotimes for the correlation between model scores and human ratings on the English {\sc Rw} dataset.
        The \mbox{csmRNNs} benefit from initialisation with high quality pre-existing word embeddings, while our models used random initialisation.
    }
    \label{tbl:wordsim_rw}
    \centering
    \ICMLpretablevskip
    \begin{small}
    %Using concatenated vectors:
    \begin{tabular}{lrlr} \toprule
        \multicolumn{2}{c}{\emph{\cite{Luong2013}}} &\multicolumn{2}{c}{\emph{Our models}} \\ \midrule
        HSMN        & 2  & \clbl       & 18 \\ 
        HSMN+csmRNN & 22 & \clblpp     & \textbf{30} \\ 
        C\&W        & 27 & $-$\compose & 20   \\ 
        C\&W+csmRNN & \textbf{34}      &   \\  
        \bottomrule
    \end{tabular}
    \end{small}
    \ICMLposttablevskip
\end{table}

\begin{figure}[bt]
    \ICMLprefigvskip
    \centering 
    \includegraphics[trim=0 1.5mm 0 1mm,clip,width=0.35\textwidth]{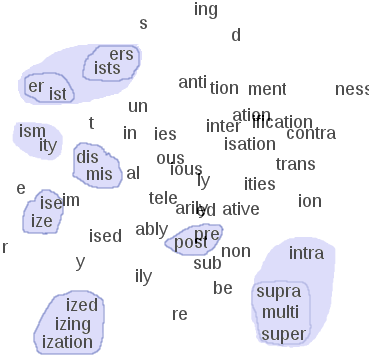}
    \caption{\textbf{English morpheme vectors learnt by \modelx{CLBL++}.} \; 
      Dimensionality reduction was performed with 
        t-SNE \cite{MaatenHintonTSNE2008}, with shading added for emphasis.
    }
    \label{fig:tsne}
    \ICMLpostfigvskip
\end{figure}

    \begin{table}[t]
        \caption{\textbf{Word-pair similarity task (multi-language),} 
             showing Spearman's \rhotimes and the number of word pairs in each dataset.
            As benchmarks, we include the best results from \newcite{Luong2013}, who relied on more training data and pre-existing embeddings not available in all languages. In the penultimate row our model's ability to compose vectors for OOV words is suppressed.
        }
        \label{tbl:wordsim_multilang}
        \ICMLpretablevskip
        \centering
        \begin{small}
        % Include csmRNN; use concatenated vectors
        \begin{tabular}{l *{6}{@{\hskip 0.1em}c} } \toprule
\qquad \quad\emph{Datasets}\footnotemark%TheDatasetReferences
            & \multicolumn{2}{c}{WS}
            & \multicolumn{1}{c}{Gur}
            & \multicolumn{2}{c}{RG}
            & \multicolumn{1}{c}{ZG}            
            \\ \cmidrule(lr){2-3} \cmidrule(lr){4-4} \cmidrule(lr){5-6} \cmidrule(lr){7-7}
Model / \emph{Language}  & \multicolumn{1}{c}{\Len}  & \multicolumn{1}{c}{\Les}
            & \multicolumn{1}{c}{\Lde}
            & \multicolumn{1}{c}{\Len}  &  \multicolumn{1}{c}{\Lfr}  &  \multicolumn{1}{c}{\Lde} \\ \midrule
            %ws_en       ws_es     %gur_de  %rg_en      &rg_fr & zg_de
            HSMN          & 63        & --       & --     & 63        & --   & -- \\ 
            +csmRNN       & 65        & --       & --     & 65        & --   & -- \\ \midrule
            \clbl       & 32        & 26       & 36     & {\bf 47}  & 33   & 6 \\
            \clblpp     & 39  & {\bf 28}    & {\bf 56}  & 41        & {\bf 45}   & {\bf 25} \\  
            $-$\compose & {\bf 40}  & 27       & 44     & 41        & 41   & 23 \\ \midrule
            \multicolumn{1}{l}{\# pairs} & \multicolumn{2}{c}{353} 
            &  \multicolumn{1}{c}{350} 
            &  \multicolumn{2}{c}{65}
            &  \multicolumn{1}{c}{222}  
            \\ \bottomrule
        \end{tabular}
        \end{small}
    \ICMLposttablevskip
    \end{table}

%TheDatasetReferences
\footnotetext{
    \Les WS353 \cite{Hassan2009}; 
    Gur350 \cite{Gurevych2005};
    RG65 \cite{RubensteinGoodenough65} with \Lfr \cite{Joubarne2011};
    ZG222 \cite{Zesch2006}.
}

A visualisation of the English morpheme vectors (\autoref{fig:tsne}) suggests the model captured non-trivial morphological regularities:
noun suffixes relating to persons (writ\emph{er}, human\emph{ists}) lie close together, while being separated according to number;
negation prefixes share a region 
(\mbox{un-}, \mbox{in-}, \mbox{mis-}, \mbox{dis-});
and relational prefixes are grouped 
(\mbox{surpa-}, \mbox{super-}, \mbox{multi-}, \mbox{intra-}),
with a potential explanation for their separation from inter- being that the latter is more strongly bound up in lexicalisations (\emph{inter}national, \emph{inter}section).

\subsection{Task 2: Machine Translation}
\label{sec:eval-mt}
The final aspect of our evaluation focuses on the integration of class-decomposed log-bilinear models into a machine translation system.
To the best of our knowledge, this is the first study to investigate large vocabulary normalised CSLMs inside a decoder when translating into a range of morphologically rich languages.
We consider 5 language pairs, translating from English into Czech, German, Russian, Spanish and French.

Aside from the choice of language pairs, this evaluation diverges from \newcite{Vaswani2013} by using normalised probabilities, a process made tractable by the class-based decomposition and caching of context-specific normaliser terms.
\newcite{Vaswani2013} relied on unnormalised model scores for efficiency, but do not report on the performance impact of this assumption.
In our preliminary experiments, there was high variance in the performance of unnormalised models. 
They are difficult to reason about as a feature function that must help the
translation model discriminate between alternative hypotheses.

We use \tool{cdec} \cite{Dyer2010, DyerFastAlign2013} to build symmetric word-alignments and extract rules for hierarchical phrase-based translation \cite{Chiang2007}.
Our baseline system uses a standard set of features in a log-linear translation model. This includes a baseline 4-gram \mkn language model,
trained with \tool{SRILM} \cite{Stolcke02srilm}
and queried efficiently using \tool{KenLM} \cite{Heafield2011}.
The CSLMs are integrated directly into the decoder as an \emph{additional} feature function, thus exercising a stronger influence on the search than in n-best list rescoring.\footnote{Our source code for using \modelx{CLBL}/\modelx{CLBL++} with \tool{cdec} is released at \ourcode.}
Translation model feature weights are tuned with MERT \cite{Och2003a} on \corpus{newstest2012}.

\begin{table}[tb]
  \caption{\textbf{Translation results.} Case-insensitive \bleu scores on \corpusx{newstest2013},
        with standard deviation over 3~runs given in parentheses.
        The two right-most columns use the listed CSLM as a feature in addition to the MKN feature, i.e.\ these MT systems have at most 2 LMs.
        Language models are from \autoref{tbl:ppl_mono} (top).
    }
    \label{tbl:bleu}
    \ICMLpretablevskip
    \centering
    \begin{small}
    \begin{tabular}{l 
            c @{}>{\small\bgroup \;(}c<{)\egroup}  
            c @{}>{\small\bgroup \;(}c<{)\egroup}
            c @{}>{\small\bgroup \;(}c<{)\egroup}
        } \toprule
        & \multicolumn{2}{c}{\mkn}   & \multicolumn{2}{c}{\clbl} & \multicolumn{2}{c}{\clblpp}  \\   
        \midrule 
        \data{En}$\rightarrow$\Lcs           & 12.6     & 0.2 & 13.2 & 0.1 & \textbf{13.6} & 0.0   \\
        \phantom{\data{En}$\rightarrow$}\Lde & 15.7     & 0.1 & \textbf{15.9} & 0.2 & 15.8 & 0.4    \\
        \phantom{\data{En}$\rightarrow$}\Les & 24.7     & 0.4 & 25.5 & 0.5 & \textbf{25.7} & 0.3    \\
        \phantom{\data{En}$\rightarrow$}\Lfr & 24.1     & 0.2 & 24.6 & 0.2 & \textbf{24.8} & 0.5    \\
        \phantom{\data{En}$\rightarrow$}\Lru & 15.9     & 0.2 & 16.9 & 0.3 & \textbf{17.1} & 0.1    \\
        \data{Cs}$\rightarrow$\Len           & 19.8     & 0.4 & \textbf{20.4} & 0.4 & \textbf{20.4} & 0.5 \\
        \bottomrule
    \end{tabular}
\end{small}
    \ICMLposttablevskip
\end{table}

\autoref{tbl:bleu} summarises our translation results.
Inclusion of the \clblpp language model feature outperforms the \mkn-only baseline systems by 1.2~\bleu points for translation into Russian, and by 1 point into Czech and Spanish.
The \lp{En}{De} 
system benefits least from the additional CSLM feature, despite the perplexity reductions achieved in the intrinsic evaluation.
In light of German's productive compounding, it is conceivable that the bilingual coverage of that system is more of a limitation than the performance of the language models.

On the other languages, the \clbl adds 0.5~to~1 \bleu points over the baseline,
whereas additional improvement from the additive representations lies within MERT variance except for \lp{En}{Cs}.

The impact of our morphology-aware language model is limited by the translation system's inability to generate unseen inflections.
A future task is thus to combine it with a system that can do so \cite{Chahuneau2013SynthPhrases}.

\section{Related Work}

Factored language models (FLMs) have been used to integrate morphological information into both discrete \ngram LMs \cite{Bilmes2003FLM} and CSLMs \cite{Alexandrescu2006} by viewing a word as a set of factors.
\newcite{Alexandrescu2006} demonstrated how factorising the representations of context-words can help deal with out-of-vocabulary words, but they did not evaluate the effect of factorising output words and did not conduct an extrinsic evaluation.

A variety of strategies have been explored for bringing CSLMs to bear on machine translation.
Rescoring lattices with a CSLM proved to be beneficial for ASR \cite{Schwenk2004} and was subsequently applied to translation \cite{Schwenk2006,SchwenkKoehn2008}, reaching training sizes of up to 500m words \cite{SchwenkGPUs2012}.
For efficiency, this line of work relied heavily on small ``shortlists'' of common words, by-passing the CSLM and using a back-off \ngram model for the remainder of the vocabulary.
Using unnormalised CSLMs during first-pass decoding has generated improvements in \bleu score for translation into English \cite{Vaswani2013}.

Recent work has moved beyond monolingual vector-space modelling, incorporating phrase similarity ratings based on bilingual word embeddings as a translation model feature \cite{ZouSocher2013}, or formulating translation purely in terms of continuous-space models \cite{NalRCTMs}.
Accounting for linguistically derived information such as 
morphology \cite{Luong2013,Lazaridou2013}
or syntax \cite{Hermann2013}
has recently proved beneficial to learning vector representations of words.
Our contribution is to create morphological awareness in a \emph{probabilistic } language model. 

\section{Conclusion}
We introduced a method for integrating morphology into probabilistic continuous-space language models.
Our method has the flexibility to be used for morphologically rich languages (MRLs) across a range of linguistic typologies.
Our empirical evaluation focused on multiple MRLs and different tasks.
The primary outcomes are that 
(i)~our morphology-guided CSLMs improve intrinsic language model performance when compared to baseline CSLMs and \ngram MKN models;
(ii)~word and morpheme representations learnt in the process compare favourably in terms of a word similarity task to a recent more complex model that used more data, while obtaining large gains on some languages;
(iii)~machine translation quality as measured by \bleu was improved consistently across six language pairs when using CSLMs during decoding, although the morphology-based representations led to further improvements beyond the level of optimiser variance only for English$\rightarrow$Czech.
By demonstrating that the class decomposition enables full integration of a normalised CSLM into a decoder, we open up many other possibilities in this active modelling space.

\begin{small}
\bibliography{tex/library}
\end{small}

\bibliographystyle{icml2014}

\end{document}